%% file: main.tex
\documentclass{article}

\usepackage{microtype}
\usepackage{graphicx}
\usepackage{subfigure}
\usepackage{booktabs} 
\usepackage{amsmath}
\usepackage{multirow}

\usepackage{hyperref}

\usepackage[accepted]{icml2019}

\icmltitlerunning{Training Question Answering Models From Synthetic Data}

\begin{document}

\twocolumn[
\icmltitle{Training Question Answering Models From Synthetic Data}



\icmlsetsymbol{equal}{*}

\begin{icmlauthorlist}
\icmlauthor{Raul Puri}{nv}
\icmlauthor{Ryan Spring}{ru}
\icmlauthor{Mostofa Patwary}{nv}
\icmlauthor{Mohammad Shoeybi}{nv}
\icmlauthor{Bryan Catanzaro}{nv}
\end{icmlauthorlist}

\icmlaffiliation{nv}{Nvidia, Santa Clara, California, USA}
\icmlaffiliation{ru}{Rice University, Houston, Texas, USA}

\icmlcorrespondingauthor{Raul Puri}{raulp@nvidia.com}
\icmlcorrespondingauthor{Ryan Spring}{rdspring1@gmail.com}

\icmlkeywords{Machine Learning, ICML}

\vskip 0.3in
]



\printAffiliationsAndNotice{}  

\newcommand{\squad}{\textsc{SQuAD }}
\newcommand{\squadone}{\textsc{SQuAD1.1 }}
\newcommand{\squadtwo}{\textsc{SQuAD2.0 }}

\begin{abstract}

Question and answer generation is a data augmentation method that aims to improve question answering (QA) models given the limited amount of human labeled data. However, a considerable gap remains between synthetic and human-generated question-answer pairs. This work aims to narrow this gap by taking advantage of large language models and explores several factors such as model size, quality of pretrained models, scale of data synthesized, and algorithmic choices. On the \squadone question answering task, we achieve higher accuracy using solely synthetic questions and answers than when using the \squadone training set questions alone. Removing access to real Wikipedia data, we synthesize questions and answers from a synthetic corpus generated by an 8.3 billion parameter GPT-2 model. With no access to human supervision and only access to other models, we are able to train state of the art question answering networks on entirely model-generated data that achieve 88.4 Exact Match (EM) and 93.9 F1 score on the \squadone dev set. We further apply our methodology to \squadtwo and show a 2.8 absolute gain on EM score compared to prior work using synthetic data.
\end{abstract}

\section{Introduction}
\begin{table}[th]
\centering\scalebox{0.835}{
\begin{tabular}{|c|p{8cm}|}
\hline
  Text
  & Albert Einstein is known for his theories of special relativity and general relativity. He also made important contributions to statistical mechanics, especially his mathematical treatment of Brownian motion, his resolution of the paradox of specific heats, and his connection of fluctuations and dissipation. Despite his reservations about its interpretation, Einstein also made contributions to quantum mechanics and, indirectly, \textbf{quantum field theory}, primarily through his theoretical studies of the photon. \\
  \hline
  117M & Which two concepts made Einstein's post on quantum mechanics relevant? \\
  \hline
  768M & Albert Einstein also made significant contributions to which field of theory? \\
  \hline
  8.3B & Because of his work with the photon, what theory did he indirectly contribute to? \\
  \hline
  Human & What theory did Einstein have reservations about? \\
  \hline
\end{tabular}}
\caption{\label{tab:example_0} Questions generated by models of increasing capacity with the ground truth answer highlighted in bold. As model size grows, question quality becomes increasingly coherent, complex, and factually relevant.}
\vspace{-4mm}
\end{table}
One of the limitations of developing models for question answering, or any Deep Learning application for that matter, is the availability and cost of labeled training data. A common approach to alleviate this need is semi-supervised learning, wherein one trains a model on existing data and uses it to label more data for training \cite{semi_lit,semi_book,semi_intro,semi_deep}. This technique has demonstrated benefits in recent literature for image classification \cite{quoc_semi} and question answering (QA) tasks \cite{roundtrip,unilm}. However, the complexities of generating questions and answers in natural language proves challenging for existing methods, with a large gap in quality remaining between synthetic and human-generated data. In this work, we close this gap using only synthetic questions generated from large models. We also show that answer candidate generation is foundational to synthetic question quality.

Consistent with prior work \cite{roundtrip,unilm}, we use a 3-step modeling pipeline consisting of unconditional answer extraction from text, question generation, and question filtration. Our approach for training question generators on labeled data uses pretrained GPT-2 decoder models and a next-token-prediction language modeling objective, trained using a concatenation of context, answer, and question tokens. As demonstrated in sections \ref{sec:qgen_scale} and \ref{sec:qgen_choices}, pretraining large generative transformer models up to 8.3B parameters improves the quality of generated questions. Additionally, we propose an overgenerate and filter approach to further improve question filtration. The quality of questions produced by this pipeline can be assessed quantitatively by finetuning QA models and evaluating results on the \squad dataset. 

We demonstrate generated questions to be comparable to supervised training with real data. For answerable \squadone questions we recover 100.4\% of fully supervised EM and F1 scores, when training on purely synthetic questions and answers generated from unlabeled data. Specifically, we achieve scores of 88.4 and 94.1 versus supervised training which achieves 87.7 EM and 94.0 F1. Finetuning the resulting model on real \squadone data reaches 89.4 EM and 95.1 F1 score, which is higher than any prior BERT-based approach. In Table \ref{tab:example_0}, we show that the generated questions are qualitatively similar to ground truth questions, with quality improving as a function of model size. 

Going further, we show that QA models can be successfully trained from fully synthetic data, by running question and answer generation on a corpus generated from an unconditional GPT-2 model. With zero access to human language supervision and only access to models, we achieve an EM of 88.4 and F1 of 93.9. This approach performs comparably to generating questions from real data recovering 100.3\% of fully supervised EM and F1 scores.

In sum, we demonstrate that recent advances in language models are capable of generating high quality answerable questions. This can be used in many ways, for example, synthesizing unanswerable questions \cite{rajpurkar2018squad2}, boolean questions \cite{boolq}, and complex questions \cite{alphazero, alphago, dota}. Outside of data augmentation and training question answering models, high quality question generation can be used to pose human-plausible questions for interpretability or generate intermediate queries for multihop reasoning. Additionally, modeling how humans ask questions can be used to improve search and information retrieval in query-conditional semantic information retrieval.
All these applications require improved question generation capability. In summary, our contributions are as follows:

\begin{itemize}
    \item We demonstrate for the first time that finetuning a model on purely synthetic questions and answers generated from a synthetic corpus, creates a QA model better in \squadone EM and F1 scores than one trained from human-labeled data.
    \item We show that by scaling the model size, using better ptretrained models, and leveraging large synthetically generated data, we achieve state of the art results and show 1.7 absolute gain on \squadtwo EM score compared to prior work using synthetic data.
    \item Through detailed ablation studies we identify that the quality of answer generation is fundamental to high fidelity question generation and properly aligning the answer distribution boosts scores by 19.8 EM points.
    
\end{itemize}

\section{Method}

In this work we seek to generate high quality training data for \squad\-style extractive question answering over a given set of documents $D$. This requires us to sample $(c, q, a)$ triples for given paragraph contexts $c\in D$ according to probability $p(q, a | c)$, where $q$ is a question resulting in answer $a$, which exists as a contiguous span of text in $c$. Leveraging the roundtrip consistency method \cite{roundtrip}, we achieve this by using a three step approach consisting of Answer Generation $\hat{a}\sim p(a|c)$, Question Generation $\hat{q}\sim p(q|\hat{a},c)$, and Roundtrip Filtration $\hat{a}\stackrel{?}{=}a^*\sim p(a|c, \hat{q})$. As illustrated by Algorithm \ref{alg:BERT_GPT_QA_GEN} the synthesized dataset of triples is then used to finetune and train a BERT-based QA model similar to \cite{bert}. 

\begin{algorithm}[!htbp]
\label{alg:BERT_GPT_QA_GEN}
\begin{algorithmic}
\STATE 1. Sample answer candidates from paragraphs using a BERT model.
\STATE 2. Generate questions from answer candidates and paragraphs using a GPT-2 model.
\STATE 3. Apply a BERT roundtrip consistency model to filter generated question answer pairs.
\STATE 4. Train a BERT QA model using filtered synthetic questions and evaluate on development set.
\end{algorithmic}
\caption{Pipeline for generating and evaluating synthetic data.}
\end{algorithm}
\vspace{-3mm}

\subsection{Answer Generation: $\hat{a}\sim p(a|c)$}
\cite{talmor-berant-2019-multiqa} empirically showed that a QA model trained on a specific dataset does not necessarily generalize well to other similar QA datasets. For a model to perform well on a specific dataset, we need to match its answer distribution. Our goal is to learn an answer candidate generator $p(a|c)$, that acts as a prior for the dataset's answer distribution. Earlier work \citep{dhingra2018simple, lewis-etal-2019-cloze} using named entity and noun phrase answer candidates performed best only on those portions of the data distribution. By aligning our answer candidates with the dataset answers through end-to-end model-based approaches, our performance is comparable to the fully-supervised baseline.

To achieve this we finetune a BERT-style transformer model with hidden size $H$ for extractive span selection. However, unlike BERT finetuning for question answering we omit the question tokens. This yields an unconditional answer extractor model $p(a|c)$ that predicts the start and end of a token span $(s,e) = a$. Similar to \cite{roundtrip} we used an answer extraction head that models start and end tokens jointly. 
\newcommand{\CtoA}{\theta_A}
   $$ p(a|c;\CtoA) =
    \frac{e^{f(a, c; \CtoA)}}
    {\sum_{a''} e^{f(a'', c; \CtoA)}} $$
    $$f (a, c; \CtoA) = \mbox{MLP}(\mbox{CONCAT}(\mbox{BERT}(c)[s], \mbox{BERT}(c)[e]))$$
We also found that joint modeling performed better than an extraction head that models start and end tokens independently. Lastly, our MLP layer consists of one hidden layer with hidden size $2H$, followed by a ReLU nonlinearity, and a projection from activations to logits.

\subsection{Question Generation: $\hat{q}\sim p(q|\hat{a},c)$}
\begin{figure}
    \centering
    \includegraphics[scale=.27]{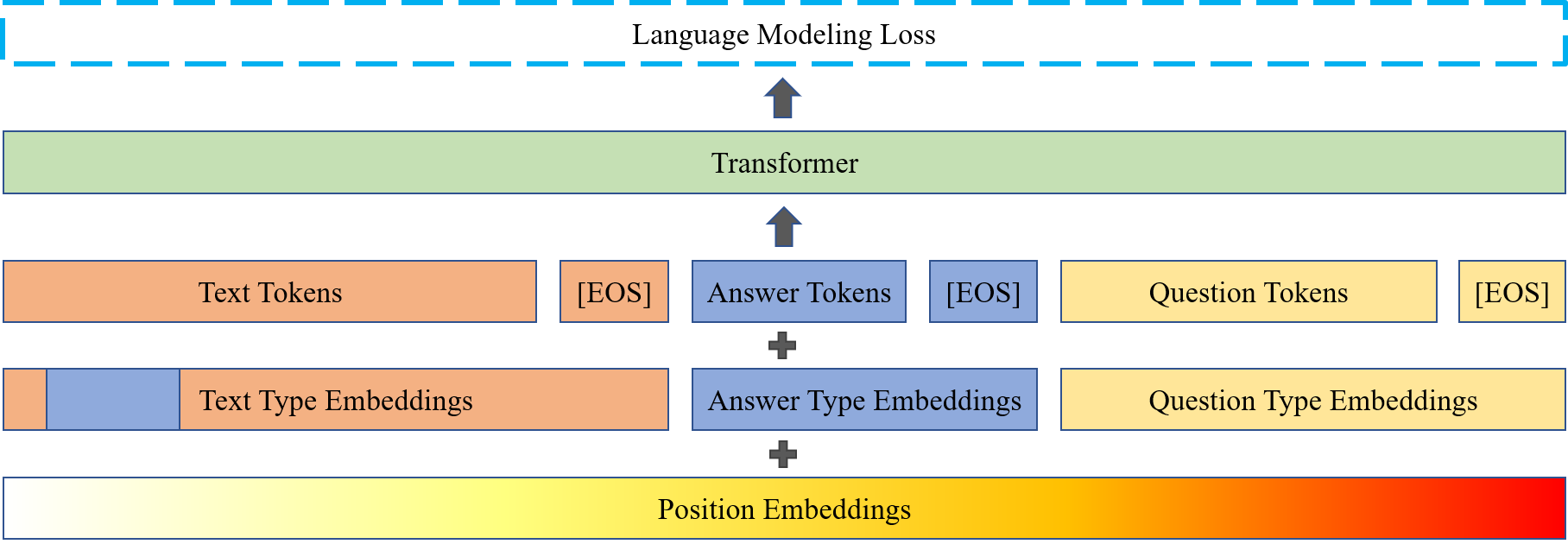}
    \vspace{-4mm}
    \caption{Question Generation input representation and language modeling loss. Answer type embeddings highlight the answer's presence in the text.}
    \label{fig:input_rep}
\vspace{-4mm}    
\end{figure}

We develop a conditional question generation model, $p(q|a,c)$ using a pretrained GPT-2 model. As input to our model, we concatenate context tokens, answer tokens, and question tokens into a single sequence, separated by the end of sequence tokens. We use three segment type embeddings to help the GPT-2 decoder model distinguish between different parts of the input. This method of multi-input controlled text generation draws on inspiration from prior work \cite{zeroshot_clf,T5,unilm}. We also use answer segment type embeddings to highlight the presence of the answer span in the provided context tokens. We trained this question generation model with a left to right next token prediction loss modeled over the entire concatenated sequence. Visualizations of the input representation and training loss can be found in Figure \ref{fig:input_rep}. To sample from our learned model we concatenate the context tokens with the answer tokens and autoregressively sample output question tokens. 

To aid our model with generation we employ start and stop word filtration. We prepend `\textit{question:}' and append `\textit{:question}' tokens to the questions in our training dataset. During inference time, if the model does not sample a sequence containing both the start and stop words we discard the example entirely. 

\subsection{Roundtrip Filtration: $\hat{a}\stackrel{?}{=}a^*\sim p(a|c, \hat{q})$}
In roundtrip filtration \cite{roundtrip} an extractive question answering model $p(a|c,q)$ is trained on the available labeled data. When a new question, answer, and context triple $(c,\hat{q},\hat{a})$ is generated we apply the QA filtration model $p(a|c, \hat{q})$ to the context and question. The resulting answer $a^*$ from the model is compared to the answer $\hat{a}$ from the triple. If the two are equivalent then the question is considered admissible. 

In the original work, however, the authors draw attention to the precision of the method. While it does discard invalid questions, several valid questions are discarded as well. To avoid losing valuable pieces of information to train our question answering models we propose generating two questions, instead of one question, for each candidate answer. Roundtrip filtration is then applied to each question individually. If a triple is decided as acceptable then it is kept regardless of whether the other triple is acceptable, leading to a scenario where both can be kept. This method is similar to prior work in overgeneration and reranking of generated questions \cite{statisticalranking}.

\section{Experiment Setup}

\begin{figure}[t!]
    \centering
    \includegraphics[scale=.55]{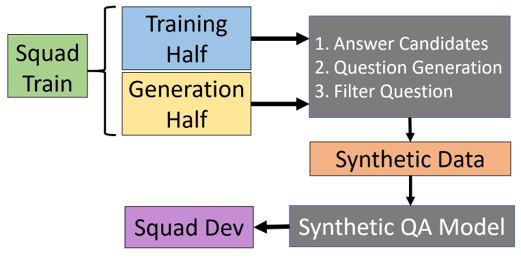}
    \vspace{-4mm}
    \caption{\label{fig:data_flow} Data flow for training and evaluating question generation pipeline. 
    }
\vspace{-4mm}    
\end{figure}

For all implementations and training of transformer models we rely on the Megatron-LM codebase \cite{Megatron}. For off-the-shelf weights and implementations of BERT-Large we rely on the HuggingFace's transformers codebase \cite{wolf2019huggingfaces}. The GPT-2 models \cite{gpt2} used for question generation were each pretrained on the 174GB corpora used in Megatron-LM: Wikipedia \cite{bert}, OpenWebText \cite{openwebtext}, RealNews \cite{realnews}, and CC-Stories \cite{ccstories}. Unless otherwise noted, our GPT-2 models were trained at a batch size of 512 for 300k iterations with 3k iterations of warmup, Adamw \cite{adamw} for optimization, a learning rate of 1.5e-4 decaying linearly to 1e-5, weight decay of 0.01, global gradient norm clipping of $1.0$, and a normal initialization of $\theta\sim\mathcal{N}(0,0.02)$. Finetuning our GPT-2 models we used the same hyperparameters except for a batch size of 32 and a learning rate of 2e-5 decaying to zero over six epochs of finetuning data. For model configurations of hidden size, number of layers, and attention heads, we used the configurations detailed in Megatron-LM.

To train our BERT models we relied on a pretraining regime similar to ALBERT. We used a n-gram masked language modeling task in conjunction with a sentence order prediction task. Unlike ALBERT we did not utilize weight sharing and we used a GPT-2 style ordering of residual connections and layer normalization. We found this greatly improved stability and allowed us to train significantly larger BERT models than prior work \cite{albert} without encountering training instabilities and overfitting. We trained our BERT models with the same hyperparameters as GPT-2 except using learning rate of 1e-4 and a batch size of 1024 over 2 million iterations with 10k iterations of warmup. Finetuning our BERT models for filtration, answer generation, filtration, and question answering was all done with a learning rate of 1e-5 and a cosine decay schedule over 2 epochs of training data. For our 1.2 billion parameter BERT model we used 24 layers, a hidden size of 2048, and 32 attention heads. We refer to our models as BERT-345M and BERT-1.2B and the original BERT model as BERT-Large.

To train and evaluate the whole question generation pipeline for our ablation studies in sections \ref{sec:model_scale} and \ref{sec:modeling_choices} we used a data partitioning scheme as detailed in Figure \ref{fig:data_flow}. A similar data pipeline has been employed in concurrent work of \cite{learn2answer_learn2ask}. We split the \squad training data into equal portions, partitioning the data randomly into two sets of documents. One half of the documents is used to train the answer generator, question generator, and filtration models while the second half of the documents is used to generate synthetic data to finetune a QA model. The finetuned QA model is then evaluated on \squad dev set, where the evaluation results are used as a surrogate measure of synthetic data quality. The partitioning of the dataset is done to avoid leakage and overfitting between the data seen at training time and generation time thereby testing the generalization capabilities of our models. Since shuffling is done randomly we repeat this process 5 times with different seeds for every ablation study and report the mean of our results. 

Lastly, all our models were trained with mixed precision training \cite{fp16training} on NVIDIA V100 GPUs. Pretraining took place on anywhere from 4 to 32 DGX-2H servers for our largest models. Finetuning only required one DGX-1V, except in the case of finetuning the 8.3B parameter question generator which required eight DGX-1Vs. 

\section{Results}
\input{generate_synthetic.tex}

\begin{figure}[t!]
    \centering
    \includegraphics[scale=.6]{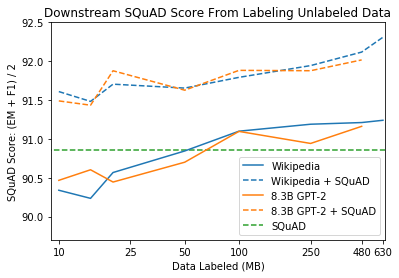}
    \vspace{-6mm}
    \caption{\label{fig:label_scale} Effect of labeling data size on downstream \squadone score. After finetuning BERT-345M models on synthetic data we finetune further on human generated \squadone data.}
\vspace{-4mm}    
\end{figure}

In this section we present our results using the best combination of models, algorithms, and parameters. In the following sections, we will perform detailed ablation study and show contributions from each of these choices. 

We train a 1.2 billion parameter answer generator, question generator, and question filtering model. In these experiments we use the entire \squadone dataset instead of only training on half of the labeled data since we are not doing any model or hyperparameter search. We then use these models to label synthetic data from two sources outside of \squadone. We first label data from real Wikipedia documents with the overlapping documents from the \squadone training and dev set removed. In parallel we label data from synthetic Wikipedia documents generated by an 8.3B GPT-2 model. This model was first trained with the Megatron-LM codebase for 400k iterations before being finetuned on only Wikipedia documents for 2k iterations. This allows us to generate high quality text from a distribution similar to Wikipedia by using top-$p$ ($p=0.96$) nucleus sampling. 

Table \ref{tab:generate_synthetic} shows results when the synthetic data is finetuned on a BERT-345M QA model. We showcase that we are able to recover and surpass the performance of real data by only using synthetic data generated from synthetic corpus. Using real questions synthesized on real Wikipedia data we do even better. Finetuning this model afterwards on the actual \squadone dataset allows us to achieve a 1.7 and 1.2 point boost to our EM and F1 scores. In Figure \ref{fig:label_scale} we examine the relationship between \squadone score and the amount of text labeled. We find that the performance of training with purely synthetic data observes a $\log$-linear relationship that begins to saturate at approximately 100 MB of text labeled. However, finetuning these models on labeled \squadone data demonstrates continued improvement even beyond saturation. The performance of these post finetuned models continues to improve even past 500 MB of data labeled. 

\input{sota_compare.tex}

\paragraph{Comparison with prior work.} To quantify the improvements in question generation quality derived from improvements to language models and our generation techniques we compare our results to the original roundtrip consistency work from \cite{roundtrip}. We generate 3 million questions from real Wikipedia text and finetune the public BERT-Large model on this data. We then finetune the model on the human-generated \squadtwo dataset and evaluate on the dev set. Unlike prior work we do not generate any unanswerable questions, yet we find in Table \ref{tab:sota_compare} that our synthetic data approach outperforms the prior work. This is despite our BERT-Large baseline underperforming the numbers reported in \cite{roundtrip} by a full point. We also compare our methods with the state of the art in synthetically trained \squadtwo \cite{unilm} and find that with a similar number of questions we outperform existing methods, and with even more labeled data this trend persists. 

\section{Model Scale}
\label{sec:model_scale}

A central premise of our work is that language models have advanced to the point where they are now able to generate diverse, coherent language and as a result can generate high quality question answering curricula. We show in this section that as we improve pretraining tasks, pretraining scale, and model scale, synthetic data also improves. To show improvements in question generation we track the resulting \squadone evaluation score when a BERT-style model is finetuned on the synthetic data. Table \ref{tab:model_scale} summarizes the benefits of using larger models for answer generation, question generation, and question filtration. The following subsections ablate this result to show the contributions from scaling individual components of the synthetic data pipeline.

\input{model_scale_tab.tex}

\subsection{Scaling Question Generation}
\label{sec:qgen_scale}

\input{qgen_scale_tab.tex}

Question generation plays a critical role in our synthetic data pipeline: it must synthesize linguistically and logically coherent text even if the text does not exist within the provided context. In this section we investigate the relationship between question generator scale and downstream \squadone performance. We isolate the quality of question generation by using ground truth answers from the \squadone dataset to generate questions and finetune a BERT model before evaluating it on the \squadone dev set. We perform no question filtration in between generation and finetuning. From our experiments in Table \ref{tab:qgen_scale} we find that \squadone performance increases monotonically. Additionally, the number of valid samples that pass stopword filtration increase with larger models, indicating bigger models maintain coherency during sampling. For comparisons with prior work we train a question answering model with our ALBERT-style BERT model (BERT-345M) and the original BERT-Large model. \cite{learn2answer_learn2ask} use a feedback loop to improve the question generator and BERT-Large question answering model. Compared to our work we find that a similarly parameterized set of models achieve equal if not better performance despite using only a single supervised pass through the data and no feedback loop.

\subsection{Scaling Answer Generation}
\label{sec:agen_scale}
\begin{table}
\centering\scalebox{0.835}{
\begin{tabular}{cccc}
\hline 
Answer Generator & \#Questions &  EM & F1 \\ 
\hline\hline
 BERT-Large &227063 & 77.7 &  87.6 \\
 BERT-345M & 229297 & 79.1 & 87.9 \\
 \textbf{BERT-1.2B} & \textbf{229067} & \textbf{79.2} & \textbf{88.3} \\
 \hline
Human Generated Answers & 42472 & 83.7 &  91.1 \\
\hline
\end{tabular}}
\caption{\label{tab:agen_scale} Comparison of answer generator pretraining and scale. 
Our 1.2 billion parameter question generator is used for generating questions.
}
\end{table}

Answer generation is equally important in our data generation pipeline. Answer generation is the first component of the pipeline and must be precise to avoid compounding errors. For answer generation we use an unconditional extractive BERT model that predicts start and end spans jointly over a given sentence. From each probability distribution we sample the entire nucleus ($p=0.9$) or the top-5 spans, choosing whichever is smaller. We arrive at this implementation based on our ablation studies in section \ref{sec:qfilter_choices}. To test the quality of the selected answers we generate questions from our 1.2 billion parameter question generator and finetune a question answering model on the synthesized questions without any filtration. In Table \ref{tab:agen_scale} we compare answer generation quality using our two trained models and the original BERT-Large model from \cite{bert}. We find that improvements in pretraining data and tasks dramatically improve answer generation quality by 1.4 EM and 0.3 F1 between BERT-Large and our 345 million parameter answer generation model. We find that increasing model scale further to 1.2 billion parameters improves answer generation quality F1 by 0.4 while EM only improves by 0.1. 
Although these represent improvements in question quality only achieved by newer models, answer generation seems to be a large bottleneck as we discuss in section \ref{sec:qgen_choices}.

\subsection{Scaling Question Filtration}
\label{sec:qfilter_scale}
\input{qfilter_scale_tab.tex}
We use the 1.2 billion parameter question generator from section \ref{sec:qgen_scale} to generate questions for filtration. As described in more detail in section \ref{sec:qfilter_choices} we overgenerate two questions for every answer. We then filter these questions with roundtrip filtration before finetuning a question answering model. In Table \ref{tab:qfilter_scale} we find that our 345 million parameter BERT model modestly outperforms the public BERT-Large model when using synthetic answers to generate questions while our 1.2 billion parameter BERT model further improves on this score by more than a whole point. Interestingly, these results follow an opposite trend compared to the previous section. In the previous section improvements to pretraining scale and tasks made a larger difference on answer generation than increasing model scale. However, here we see the opposite: improvements to pretraining tasks results only in a modest improvement to question filtration, while increasing model size results in much more substantive improvements. We hypothesize that this is due to the larger model's ability to correctly answer more questions, and therefore allow more valid and high quality samples through to the finetuning phase as indicated by the number of questions generated by the technique.

\section{Modeling Choices}
\label{sec:modeling_choices}
While developing our synthetic data generation pipeline we explored several modeling and algorithmic choices before scaling up the model size and data quantity used. We pursued three axis of investigation, ablating choices for each model component of our pipeline at a time. While this analysis does not capture second-order effects that may arise from combining different hyperparameters across studies, we believe that it captures general trends within the space of possible modeling options.

\subsection{Question Generation}
\label{sec:qgen_choices}
\input{qgen_choices_tab.tex}
To study question generation in isolation we used our 345 million parameter model to generate questions from ground truth \squadone answers. The results of our analysis can be found in Table \ref{tab:qgen_choices}. We first investigated the use of pretrained models and found that pretraining our GPT-2 model was crucial for achieving reasonable question generation. We then examined the effect of stopword filtration in our question generator. We found that this provided a substantial boost to EM and F1 scores of 5.2 and 4.1 respectively. The goal of employing this technique was to catch generations that ramble onwards without stopping, or produce end of text prematurely in the middle of a question. On manual inspection we found qualitatively that this technique helped when generating questions on text that featured heavy use of symbols and foreign language. In these cases the model struggled with out of distribution vocabulary, autoregressive sampling degenerated, and no stopword was produced. 

\subsection{Answer Generation}
\label{sec:agen_choices}
\input{agen_choices_tab.tex}
\begin{figure}[t!]
    \centering
    \includegraphics[scale=.62]{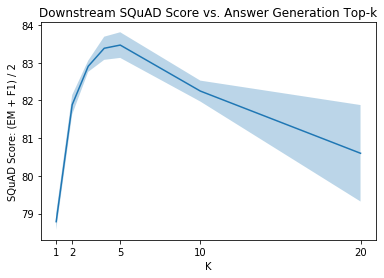}
    \vspace{-6mm}
    \caption{\label{fig:agen_topk} Effect of top-$k$ answer generation on downstream \squadone performance. For a particular value of $k$ we sample all top-$k$ candidate answers (within a nucleus of $p=0.9$) from a sequence according to a 345M parameter answer generator.}
\vspace{-3mm}
\end{figure}
In our experiments we found answer generation to be a significant bottleneck in performance. In section \ref{sec:agen_scale} we found that scaling up model size allows us to close the gap between human and synthetic training performance. However, these scaling analysis were performed with our best model. In Table \ref{tab:agen_choices} we show that the choice of model is critical to closing the gap. Starting with a Named Entity Recognition (NER) model we find that it gets a dismal EM and F1 score. This is due to entities comprising only of $\sim$ 50\% of the answer distribution for \squadone. It's necessary to use a learned model to model the diverse set of answers present \squadone. We then tried to use the most common \squadone model, which models the start and end of a span independently, to extract answers from individual sentences. This performed noticeably better, boosting our score to 77.2 EM, despite producing fewer answers than NER extraction. However, upon inspection we found that modeling the span independently resulted in sampling repetitive answers. We then tried using the answer generator from \cite{roundtrip} which models the start and end of a span jointly as a conditional random field. This is the model we ended up choosing as it performed the best with an exact match score of 79.1. Lastly, we also considered jointly modeling answer spans over an entire paragraph instead of a single sentence. However, we found that it performed worse than independent span modeling over sentences.

When sampling our answer candidates we used all top-$k$ answers comprising of the top-$p$ ($p = 0.9$) nucleus of the distribution. We performed an ablation study to select $k$ as we found that this had a noticeable impact on downstream accuracy. In Figure \ref{fig:agen_topk} we found that there was an optimal spot of $k=5$ answers per sentence. When generating answers sampled from an entire paragraph we used $k=24$ as we found that there were 4.86 sentences per paragraph on average. In general, answer generation proves to be a bottleneck in our question generation pipeline. 
The difficulty in answer generation is that not only must the answers be useful and well-formed, one must solve a one-to-many modeling problem to sample multiple answers from one passage. We believe that this might also be a contributing factor behind the poorer performance of the paragraph-level answer generation.

\subsection{Question Filtration}
\label{sec:qfilter_choices}

\input{qfilter_choices_tab.tex}
Both question generation and answer generation sometimes produces poor answers. As we show in Table \ref{tab:qfilter_choices} generating synthetic data from synthetic answers without filtering deteriorates significantly, while roundtrip consistency combats this effect to perform 7.2 EM points better. However, we find that even on questions generated from ground truth answers roundtrip filtering throws away questions associated with perfectly good answers. Throwing away data significantly hurts BERT-Large whose pretrained features are not as robust as our BERT-345M model and require more finetuning data. To combat this we take an approach similar to overgeneration and reranking \cite{statisticalranking} where we generate two questions per answer and feed each into roundtrip filtration independently. We term this overgeneration and filtration. This helps avoid losing important answers in our synthesized training set. To perform overgeneration we sample one question with top-$k$ ($k=40$) sampling and one with top-$p$ ($p=0.9$) nucleus sampling. This leads approximately to a whole point of improvement for our model in both the case with and without ground truth answers, and allows us to surpass training with real \squadone data. 

\section{Related Work}
The concept of generating questions for low-resource domains and data augmentation is not new. Early work using rule based question generation \cite{heilman2010ranking} proposed the idea of over-generating and re-ranking questions with regression models learned over handcrafted linguistic features. With the proliferation of deep neural networks in Natural Language Processing (NLP), question generation advanced to use learned LSTM models \cite{du_cardie} on extractive question answering datasets such as \squad. These early works focused primarily on generating questions without explicit extracted answers in the text. However, recent, rapid improvements in language modeling and text generation \cite{devlin2019bert, gpt2} have opened the possibility of generating realistic, high-quality answer-aware questions. Answer aware question generation conditions a generative model on context, and a selected answer from the context to generate a question. The current state of the art leverages transformer based language modeling including \cite{roundtrip,unilm,learnunanswerable, learn2answer_learn2ask}.

\cite{roundtrip} uses seq2seq models to generate questions, and then enforce answer consistency on synthetic questions to filter out poorly generatesd questions in a technique called roundtrip consistency. \cite{unilm} use a unified transformer rather than a seq2seq model to generate QA data in conjunction with roundtrip consistency. They also develop a rule based method for synthesizing unanswerable questions from generated questions. \cite{learnunanswerable} go a step further to learn a model that can generate unanswerable questions from a given answerable example. 

The process of generating answers for answer-aware question generation in recent literature has primarily leveraged cloze fill-in-the-blank passages to highlight an answer in a given context. Some work uses NER or linguistic parsers to select passages for cloze translation as in  \cite{lewis-etal-2019-cloze,dhingra2018simple}. These methods are only able to generate answers for a subset of questions as \squadone is only made up of 52\% Named Entity Answers. More recent work such as \cite{roundtrip, unilm} use model based approaches to match the answer distribution of QA datasets and extract more complex answers.

However, prior work is still lacking in data quality as the resulting QA models are far from the top of the leaderboard. To improve the quality of synthetic data generation and downstream QA models, improving language model quality is crucial. In addition to pretraining task innovation, BERT \cite{bert}, RoBERTa \cite{roberta} and ALBERT \cite{albert} have showed that increasing the size of available pretraining data directly improves downstream discriminative task performance. T5 \cite{T5}, GPT-2 \cite{gpt2}, CTRL \cite{CTRL}, Megatron-LM \cite{Megatron}, and \cite{zeroshot_clf} have shown that increasing language model scale improves the quality, coherency, and correctness of text generation. The models used in \cite{T5,CTRL,gpt2,zeroshot_clf} also demonstrate that larger models allow for better control in conditional language generation. Answer generation and question generation require improvements to discriminative and conditional generative modeling, respectively. 

\squad\-style extractive question answering is not the only form of question answering. There are many other datasets covering a wide range of QA such as multihop \cite{hotpot, wikihop, cwq}, Yes-No question \cite{boolq}, trivia questions \cite{triviaqa, searchqa}, analytical questions \cite{DROP}, conversational and generative QAs \cite{coqa}, unanswerable questions \cite{rajpurkar2018squad2, alberti2019nq}, and large multitask question answering datasets \cite{talmor-berant-2019-multiqa}. While these are outside the scope of the current work, the insights developed improving quality for extractive \squad questions will aid in generating high quality questions for other datasets.

\section{Conclusion}
We build upon existing work in large scale language modeling and question generation to push the quality of synthetic question generation. With our best models, we generate large question answering datasets from unlabeled Wikipedia documents and finetune a 345 million parameter BERT-style model achieving 88.4 EM score. Finetuning the resulting model on real \squadone data further boosts the EM score to 89.4.
This amounts to a 1.7 point improvement over our fully supervised baseline. Finally, we generate synthetic text from a Wikipedia-finetuned GPT-2 model, generate answer candidates and synthetic questions based on those answers, and then train a BERT-Large model to achieve similar question answering accuracy without directly using any real data at all. Doing so required us to scale model size for our answer generators, question generators, and filtration models. 
We hope that better synthetic questions will enable new breakthroughs in question answering systems and related natural language tasks.

\bibliography{bibliography}
\bibliographystyle{icml2019}

\clearpage

\appendix
\newcommand{\abf}[1]{{\color{red} \textbf{#1}}}
\section{Samples Generated from Wikipedia Documents}
Below are synthetic question and answering pairs synthesized from real Wikipedia documents. Question and answer generation and filtration were performed by 1.2 billion parameter models finetuned over the entire \squadone dataset. Generated answer spans are bolded in the text.

\fbox{%
\parbox{0.47\textwidth}{%
\hrule \vskip 0.1in
{\bf Question}: What indicates there must be data deletion early on in the visual pathway?

\vskip 0.1in\hrule \vskip 0.1in
{\bf Context}: Evidence suggests that our visual processing system engages in bottom-up selection. For example, \abf{inattentional blindness} suggests that there must be data deletion early on in the visual pathway. This bottom-up approach allows us to respond to unexpected and salient events more quickly and is often directed by attentional selection. This also gives our visual system the property of being goal-directed. Many have suggested that the visual system is able to work efficiently by breaking images down into distinct components. Additionally, it has been argued that the visual system takes advantage of redundancies in inputs in order to transmit as much information as possible while using the fewest resources.}}
\fbox{%
\parbox{0.47\textwidth}{%
\hrule \vskip 0.1in
{\bf Question}: What type of antibiotic is cefalotin?

 \vskip 0.1in\hrule  \vskip 0.1in
{\bf Context}: Cefalotin (INN) or cephalothin (USAN) is a \abf{first-generation cephalosporin} antibiotic. It was the first cephalosporin marketed (1964) and continues to be widely used. It is an intravenously administered agent with a similar antimicrobial spectrum to cefazolin and the oral agent cefalexin. Cefalotin sodium is marketed as Keflin (Lilly) and under other trade names.}}
\fbox{%
\parbox{0.47\textwidth}{%
\hrule \vskip 0.1in
{\bf Question}: What did ``Wanted Dead or Alive" rank on the Billboard Hot 100?

 \vskip 0.1in\hrule  \vskip 0.1in
{\bf Context}: ``Wanted Dead or Alive" is a song by American rock band Bon Jovi. It is from their 1986 album "Slippery When Wet". The song was written by Jon Bon Jovi and Richie Sambora and was released in 1987 as the album's third single. During a February 20, 2008 encore performance in Detroit, Jon Bon Jovi told the crowd about running into Bob Seger at a Pistons game. As he introduced his song ``Wanted Dead or Alive", he said it was inspired by Seger's ``Turn the Page" hit and called the song the band's anthem. The song peaked at \abf{\#7} on the ``Billboard" Hot 100 chart and \#13 on the Mainstream Rock Tracks chart, making it the third single from the album to reach the Top 10 of the Hot 100. As a result, ``Slippery When Wet" was the first hard rock/glam metal album to have 3 top 10 hits on the ``Billboard" Hot 100.}}
\fbox{%
\parbox{0.47\textwidth}{%
\hrule  \vskip 0.1in
{\bf Question}: Who played the role of Othello in the scene?

 \vskip 0.1in\hrule  \vskip 0.1in
{\bf Context}: The book begins when Kostya and his fellow students are waiting for their first lesson with the Director. They are excited and nervous at the prospect of meeting, and are surprised when he tells them that their first exercise is to put on a few scenes from a play. Kostya and two of his friends perform scenes from ``Othello", with \abf{Kostya} taking the leading role. Afterwards the Director tells them their mistakes.}}
\fbox{%
\parbox{0.47\textwidth}{%
\hrule  \vskip 0.1in
{\bf Question}: Who was Miss United Kingdom in 1997?

 \vskip 0.1in\hrule  \vskip 0.1in
{\bf Context}: \abf{Vicki-Lee Walberg} (born 11 October 1975) is a model who was Miss United Kingdom in 1997, and made the top 10 at the Miss World 1997 pageant. She was the last title holder to advance to the semifinal of the contest. Walberg later went on to work in television and was a `Dolly Dealer' in Bruce Forsyth's Play Your Cards Right on ITV during its 2002 revival.}}
\fbox{%
\parbox{0.47\textwidth}{%
\hrule  \vskip 0.1in
{\bf Question}: Who broke the Phantom's mind?

 \vskip 0.1in\hrule  \vskip 0.1in
{\bf Context}: In the final episode of the game, it is revealed that Fulbright is in fact deceased, and that the Fulbright seen throughout the game is an international spy known as the Phantom posing as him, as well as the one behind most of the game's major events. Seven years prior to the game's events, the Phantom was the catalyst of the UR-1 Incident, having murdered Metis Cykes, Athena's mother, sabotaged the HAT-1 shuttle, and leaving Simon Blackquill to take the fall for the crime after seemingly incriminating evidence was found to point to Simon as the only suspect. Simon willingly allowed himself to be imprisoned in order to protect Athena and to draw the Phantom out, but Athena suffered severe trauma from the ordeal, having believed for 7 years that she had actually murdered her mother, when in fact she stabbed the Phantom in the hand in self-defense. In the present day, the Phantom attempted to finish their case, murdering Clay Terran and bombing both the HAT-2 shuttle and a courtroom in a desperate attempt to destroy incriminating evidence from the UR-1 incident. The Phantom possesses a unique psychological makeup, showing very little, if any, emotion of any sort, nor any fear. The Phantom also has no sense of self, claiming they do not know what their original gender, face, nationality, or identity even was in the beginning; having taken on so many disguises and identities, the Phantom is an endless void. However, \abf{Phoenix, Apollo, and Athena} eventually managed to break the emotionless Phantom severely in court, causing them to suffer a severe identity crisis, moments before an unseen sniper rifle takes the Phantom's life.}}
\fbox{%
\parbox{0.47\textwidth}{%
\hrule  \vskip 0.1in
{\bf Question}: What was the final score for the Tottenham home match against Newcastle United?

 \vskip 0.1in\hrule  \vskip 0.1in
{\bf Context}: He scored his first Premier League hat-trick in a 4-0 away win on Boxing Day against Aston Villa. On 5 January 2013, Bale scored in the FA Cup third round fixture against Coventry City as well as assisting Clint Dempsey on both of his goals in a 3-0 win. On 30 January, Bale scored a magnificent solo effort in the 1-1 draw with Norwich City. Bale then scored against West Bromwich Albion in a 1-0 away win on 3 February. Bale then took his goal tally of the season to 15 goals with a brace against Newcastle United in a match which Spurs won \abf{2-1}. This took Spurs into third place, and strengthened their Champions League ambitions.}}
\fbox{%
\parbox{0.47\textwidth}{%
\hrule  \vskip 0.1in
{\bf Question}: Who was arrested along with Ernst Sekunna?

 \vskip 0.1in\hrule  \vskip 0.1in
{\bf Context}: The arrests started in March 1917, with \abf{Chandra Kanta Chakraverty} ``a thin-faced, falsetto-voiced Hindu, a native of Bengal, and a speaker of many languages", and the German, Ernst Sekunna, being arrested on charges of conspiracy. Most of the others were arrested on April 8, including Franz Bopp, the German Consul General for San Francisco, E. H. von Schack, Deus Dekker and Wilhelm von Brincken. The Indian Nationalists were accused of taking ``advantage of American neutrality to plot on American soil against the allies" at ``the expense of the laws and hospitality of the United States". The two men had also taken out trade names to do business as ``The Oriental Society", ``The Oriental Kitchen",and the ``Oriental Review", and purchased of land in an isolated part of New York State.}}
\fbox{%
\parbox{0.47\textwidth}{%
\hrule  \vskip 0.1in
{\bf Question}: What protected the hulls of the Chiyoda?

 \vskip 0.1in\hrule  \vskip 0.1in
{\bf Context}: ``Chiyoda" was a belted cruiser based on a much scaled-down version of the Royal Navy’s. The hull was made of 84 watertight compartments, protected with \abf{Harvey armor}. Originally designed to carry 12.6 inch Canet guns, the plan was abandoned due to excessive top weight. Instead, the design was changed so that her main battery consisted of ten QF 4.7 inch /40 naval guns in single mounts, mounted one each in the bow and stern, and four on each side in sponsons. The use of the Elwick quick-firing technology resulted in an increase in the rate of fire by six-fold over previous cruiser designs. Her secondary battery consisted of 14 QF 3 pounder Hotchkiss and three 11-mm, 10-barrel Nordenfelt guns. She was also equipped with three Whitehead torpedo tubes mounted on the main deck. As was standard practice at the time, the prow was reinforced for ramming.}}

\section{Samples Generated from GPT-2 Documents}
Below are synthetic question and answering pairs synthesized from fake Wikipedia documents sampled unconditionally from an 8.3B GPT-2 model. Question and answer generation and filtration were performed by 1.2 billion parameter models finetuned over the entire \squadone dataset. Generated answer spans are bolded in the text.
\fbox{%
\parbox{0.47\textwidth}{%
\hrule  \vskip 0.1in
{\bf Question}: What is a clique in a DAG?

 \vskip 0.1in\hrule  \vskip 0.1in
{\bf Context}: The main purpose of the conjecture is to quantify the perfect matchings of the vertices of a graph, in a way that can be related to the number of cliques. A perfect match of two vertices means that if the graph is ``cut along the line segment connecting these two vertices", then the pair of vertices forms an optimal matching. A clique is \abf{a small subgraph} that contains all but one pair of vertices in the graph and so these perfect matchings form an ``array" of cliques with the same size as the original graph, and thus can be described by the same number of cliques.}}
\fbox{%
\parbox{0.47\textwidth}{%
\hrule  \vskip 0.1in
{\bf Question}: What property is the difference between Bis(diphenylphosphino)methane and benz(diphenylphosphino)methane?

 \vskip 0.1in\hrule  \vskip 0.1in
{\bf Context}: Bis(diphenylphosphino)methane has been found to be a \abf{sterically hindered} isomer of benz(diphenylphosphino)methane (CHPH) and therefore it has an oxidation number of 1.}}
\fbox{%
\parbox{0.47\textwidth}{%
\hrule  \vskip 0.1in
{\bf Question}: Who was in charge of the SOE during World War II?

 \vskip 0.1in\hrule  \vskip 0.1in
{\bf Context}: By 1939, the Republican cause was being supported by both the Soviet Union and the Third Reich. The SOE, led by \abf{Colonel Hugh Sinclair}, had been active in the country since 1934, delivering weapons and propaganda material to the Republicans via agents such as future French Resistance leader Francois de La Rocque. This work came to an abrupt end in April 1939, when the Germans invaded the country. Sinclair organised a flight to France, but only about a dozen agents and journalists escaped from the country.}}
\fbox{%
\parbox{0.47\textwidth}{%
\hrule  \vskip 0.1in
{\bf Question}: When did Henry II invade Normandy?

 \vskip 0.1in\hrule  \vskip 0.1in
{\bf Context}: During the reign of Louis VII of France, Eleanor was awarded by her husband the County of Anjou. In \abf{1157}, Henry II of England invaded Normandy to take possession of that duchy, defeating Louis's troops in the Battle of Brémule. Louis's grandson and heir, William Adelin, left Anjou for his home in the south of France, where he was crowned at Toulouse on 24 April 1158.}}
\fbox{%
\parbox{0.47\textwidth}{%
\hrule  \vskip 0.1in
{\bf Question}: What does Dick Grayson use as his name?

 \vskip 0.1in\hrule  \vskip 0.1in
{\bf Context}: Meanwhile, on his return to the fifth dimension, the leader of the Faceless Ones is killed in the ensuing battle and his daughter is captured. She asks the Faceless Ones for an escape plan and is told that she must first find her father's "labyrinth". The Faceless Ones then freeze her in time and her journey begins. Batman, now imprisoned in Arkham Asylum is visited by Dick Grayson in his new identity of \abf{Nightwing}. Nightwing informs him that he has broken his parole and is now hunting him. Batman is shocked to discover that Nightwing has come to Arkham because of a deal he made with the Riddler to help him track down some of Batman's other enemies. Batman is sent by the Joker to assist Nightwing, Deadman, Deathstroke, and Lex Luthor, in tracking down Deadman's apparent killer. Batman eventually learns that the person who really killed Deadman was his fellow Justice League member, Zauriel. Zauriel is revealed to be a deeply troubled angel-like figure who blames the world for the suffering and death that he has witnessed as he has been with Batman since the death of Damian Wayne. The story arc culminated in a battle in the House of Mystery between the Spectre and Zauriel in an attempt to bring the demon back to Hell. In the end, Batman accepts Zauriel's invitation to follow him back to the fifth dimension to spare him any further pain and humiliation.}}
\fbox{%
\parbox{0.47\textwidth}{%
\hrule  \vskip 0.1in
{\bf Question}: Who do Jim, Pam, Dwight, Oscar, and Jim's father, Henry attend the wedding reception for?

 \vskip 0.1in\hrule  \vskip 0.1in
{\bf Context}: At the photo shoot, Andy Bernard (Ed Helms) and Erin Hannon (Ellie Kemper) go on a fake zombie honeymoon in the office, having an intimate moment that is interrupted when they encounter a horde of the undead. Michael and Dwight then stop the zombies from approaching Andy and Erin and create a barricade. The horde is scared off, but the building must be sealed off because the zombies have damaged the power generator, resulting in a total loss of power. After the power returns, Jim, Pam, Dwight, Oscar, and Jim's father, Henry (Brock Peters), begin gathering their families and friends to go to \abf{Erin and Andy}'s wedding reception in the Scranton branch's conference room.}}
\fbox{%
\parbox{0.47\textwidth}{%
\hrule  \vskip 0.1in
{\bf Question}: What was the title of 50 Cent's first album?

 \vskip 0.1in\hrule  \vskip 0.1in
{\bf Context}: ``I Got Mine" is a song by American rapper 50 Cent from his debut studio album ``\abf{Get Rich or Die Tryin'}" (2003). The song features a guest appearance from fellow New York City rapper Nas, who was also featured on the previous single from ``Get Rich or Die Tryin'", ``Hate Me Now".}}
\fbox{%
\parbox{0.47\textwidth}{%
\hrule  \vskip 0.1in
{\bf Question}: What happens to a star when it bursts into a thermonuclear reaction?

 \vskip 0.1in\hrule  \vskip 0.1in
{\bf Context}: When the star explodes, the material is compressed to several hundred times its original size, igniting a thermonuclear reaction. This reaction causes the star to \abf{explode outward}. The first stage of the supernova explosion is not yet far enough away to reach this red giant stage, so the star is engulfed in a supernova explosion. As the star is heated up by the supernova explosion, the outer layers of the star collapse. The compression that occurred when the shock wave reached the star's surface begins to occur at the point where the star's surface meets its core. This core-surface compression heats up and accelerates the remaining core material, producing a shock wave that expands out from the core.}}
\fbox{%
\parbox{0.47\textwidth}{%
\hrule  \vskip 0.1in
{\bf Question}: What style was used in This Wonderful Life's production?

 \vskip 0.1in\hrule  \vskip 0.1in
{\bf Context}: In 2009, Maine College of Art (main campus) presented ``This Wonderful Life" as the kick-off production to their 2009/2010 theater season. Director Todd Ziegler created a minimalist approach to the production, relying mostly on the basic premise and atmosphere of the film to create a world. The Main Stage theater was transformed into an \abf{Art Deco}-esque set with minimal set pieces, provided by Redlich + Feuer Design. This setting was contrasted by the minimalistic approach to lighting, provided by Brian Claypool, that lent the production a somber tone. In keeping with the Art Deco styling, costume design and construction was done entirely by students of the Department of Theater and Dance. The music was provided by the joint choirs of the college and the Maine All State Honor Choir.}}
\fbox{%
\parbox{0.47\textwidth}{%
\hrule  \vskip 0.1in
{\bf Question}: Which road through the Texas scrublands is a controlled access road?

 \vskip 0.1in\hrule  \vskip 0.1in
{\bf Context}: The western terminus of US 83 is located on the southeast corner of the Texas-New Mexico border at the Van Horn, Texas-Van Horn, Texas city limit line. From the border the highway follows Texas State Highway 116, which crosses US 87 in Van Horn and overlaps US 70. US 83 then crosses US 87 again near Marfa, intersecting US 87 Business and Texas State Highway 292. US 83 continues west from Marfa along Highway 290, a route now called the Trans-Pecos Highway. While \abf{US 290} is a controlled-access road, it still has a large number of at-grade intersections, due to the rugged terrain. Between Marfa and Valentine, US 83 travels through the Texas scrubland of the Big Bend.}}

\end{document}

%% file: generate_synthetic.tex
\begin{table}
\centering\scalebox{0.835}{
\begin{tabular}{cccccc}
\hline 
Text                        & Source                  & finetune   & \multirow{2}{*}{\# Questions} & \multirow{2}{*}{EM} & \multirow{2}{*}{F1}   \\
Source                      & Data Size               & data       &              &               &               \\ \hline\hline
\multirow{2}{*}{Wikipedia}  & \multirow{2}{*}{638 MB} & Synthetic  &  19,925,130  & 88.4          & 94.1          \\ 
                            &                         & +\squad    &  20,012,729  & \textbf{89.4} & \textbf{95.2} \\ \hline
\multirow{2}{*}{8.3B GPT-2} & \multirow{2}{*}{480 MB} & Synthetic  &  17,400,016  & 88.4          & 93.9          \\ 
                            &                         & +\squad    &  17,487,615  & \textbf{89.1} & \textbf{94.9} \\ \hline
\squadone &     14MB      & \squad     &  87,599      & 87.7          & 94.0 \\ \hline
\end{tabular}}
\caption{\label{tab:generate_synthetic} Finetuning BERT-345M on synthetic and human-generated data. Using 1.2B parameter models we synthesize question answer pairs from real Wikipedia corpus and synthetic cospus generated from an 8.3B GPT-2 model. Completely synthetic data does better than training with real data. Finetuning with real \squadone data afterwards further boosts performance. }
\end{table}

%% file: sota_compare.tex
\begin{table}
\centering\scalebox{0.835}{
\begin{tabular}{lcc}
\hline 
Implementation &  EM & F1 \\ 
\hline\hline
 BERT-Large \cite{roundtrip} & 78.7 & 81.9 \\
 ~~~+ 3M Questions & 80.1 & 82.8 \\
 \hline
 UniLM \cite{unilm} &  80.5 & 83.4 \\
 ~~~+  9M Questions & 84.7 & 87.6 \\
\hline
 BERT-Large & 77.4 & 80.6 \\
 ~~~+  3M Questions & 81.6 & 84.5 \\
 \hline
 BERT-345M & 84.9 & 88.2 \\
 ~~~+  3M Questions & 85.8 & 88.6 \\
 \textbf{~~~+  8M Questions} & \textbf{86.4} & \textbf{89.2} \\
 \hline
 
\end{tabular}}
\caption{\label{tab:sota_compare} Comparison with prior work. 
Improvements in question generation allow for improved \squadtwo score even without generating unanswerable questions.
}
\vspace{-4mm}
\end{table}

%% file: model_scale_tab.tex
\begin{table}[!htbp]
\centering\scalebox{0.835}{
\begin{tabular}{cccc|ccc}
\hline 
\multicolumn{4}{c|}{Model Size} & \multirow{2}{*}{\# Questions} &
\multirow{2}{*}{EM} & \multirow{2}{*}{F1} \\
\cline{1-4}
Answer & Question & Filter & QA & & &  \\ 
\hline\hline
 345M & 345M  & 345M & 345M & 116721 & 85.3 & 92.0 \\
 \textbf{1.2B} & \textbf{1.2B}  & \textbf{1.2B} & \textbf{345M} & \textbf{184992} & \textbf{87.1} & \textbf{93.2} \\
\hline
\multicolumn{3}{c}{Human Generated Data} & 345M & 42472 & 86.3  & 93.2 \\
\hline
\end{tabular}}
\caption{\label{tab:model_scale} \squadone performance using synthetic data. Downstream QA models used in all experiments are 345M parameters.}
\end{table}

%% file: qgen_scale_tab.tex
\begin{table}[!htbp]
\centering\scalebox{0.835}{
\begin{tabular}{cccc}
\hline 
Question Generator & \# Questions & EM & F1 \\ 
\hline\hline
 117M & 42345 & 76.6 & 85.0 \\
 345M \cite{learn2answer_learn2ask} & - & 75.4 &  84.4 \\
 345M (w/ BERT QA model) & 42414 & 76.6 & 84.8 \\
 345M & 42414 & 80.7 & 88.6 \\
 768M & 42465 & 81.0 &  89.0 \\
 1.2B & 42472 &  83.4 & 90.9 \\
 \textbf{8.3B} & \textbf{42478} & \textbf{84.9} & \textbf{92.0} \\
 \hline
Human Generated Data & 42472 &  86.3  & 93.2 \\
\hline
\end{tabular}}
\caption{\label{tab:qgen_scale} Effect of question generator scale on \squadone performance. Ground truth answers are used to generate questions without filtration for finetuning.}
\end{table}

%% file: qfilter_scale_tab.tex
\begin{table}
\centering\scalebox{0.835}{
\begin{tabular}{cccc}
\hline 
Filter Model & \# Questions & EM & F1 \\ 
\hline\hline
\multicolumn{4}{c}{Synthetic Questions + Real Answers}\\
\hline
 BERT-Large & 45888 & 84.5 &  91.4 \\
 BERT-345M & 34341 & 84.2 & 91.4 \\
 \textbf{BERT-1.2B} & \textbf{47772} & \textbf{85.6} & \textbf{92.4} \\
 \hline
 \multicolumn{4}{c}{Synthetic Questions + Synthetic Answers}\\\hline
 BERT-Large & 177712 & 85.5 &  91.9 \\
 BERT-345M & 144322 & 85.9 & 92.5 \\
 \textbf{BERT-1.2B} & \textbf{184992} & \textbf{87.1} & \textbf{93.2} \\
 \hline
Human Generated Data & 42472 & 86.3 & 93.2 \\
\hline
\end{tabular}}
\caption{\label{tab:qfilter_scale} Effect of pretraining and scale on question filtration. Synthetic questions and answers were both generated with 1.2 billion parameter models. Before finetuning, overgeneration and filtration were performed with the models ablated here.
}
\end{table}

%% file: qgen_choices_tab.tex
\begin{table}
\centering\scalebox{0.835}{
\begin{tabular}{cccc}
\hline 
Question Generator & \# Questions & EM & F1 \\ 
\hline\hline
 \textbf{345M} & \textbf{42414} & \textbf{80.7} &  \textbf{88.6} \\
 345M (no pretraining) & {42408} & 42.7 & 51.4 \\
 345M (no stopwords) & {42486} & 75.5 & 84.5 \\
 \hline
Human Generated Questions & 42472 & 86.3 & 93.2 \\
\hline
\end{tabular}}
\caption{\label{tab:qgen_choices} Effect of question generator modeling choices. Questions are generated from ground truth answers without any filtration.}
\vspace{-4mm}
\end{table}

%% file: agen_choices_tab.tex
\begin{table}[t!]
\centering\scalebox{0.835}{
\begin{tabular}{cccc}
\hline 
Answer Generator & \# Questions & EM & F1 \\ 
\hline\hline
 NER & 132729 & 59.3 &  70.5 \\
 Independent Spans & 83534 & 77.2 & 87.1 \\
 \textbf{Joint Spans} & \textbf{229297} & \textbf{79.1} & \textbf{87.9} \\
 Paragraph-level Joint Spans & 226672 & 77.3 & 86.9 \\
 \hline
Human Generated Answers & 42472 & 83.7 &  91.1 \\
\hline
\end{tabular}}
\caption{\label{tab:agen_choices} Effect of answer generator modeling choices. Model based answer generation is performed with BERT-345M and questions are generated using a 1.2B parameter model. No filtration is applied to the generated questions.}
\end{table}

%% file: qfilter_choices_tab.tex
\begin{table}
\centering\scalebox{0.835}{
\begin{tabular}{c|c|cc|cc}
\hline 
\multirow{2}{*}{Filter Model} & \multirow{2}{*}{\# Questions} & \multicolumn{2}{|c|}{345M QA} & \multicolumn{2}{|c}{Large QA}\\
\cline{3-6}
 & & EM & F1 & EM & F1 \\ 
\hline\hline\multicolumn{6}{c}{Synthetic Questions + Real Answers}\\\hline
 None & 42472 & 83.4 & 90.9 & 79.0 & 87.0 \\
 Roundtrip (RT) & 24310 & 84.0 & 91.3 & 76.5 & 84.4  \\
 \textbf{Overgenerate \& RT} & \textbf{47772}  & \textbf{85.6} & \textbf{92.4} & \textbf{81.7} & \textbf{88.7} \\
 \hline\multicolumn{6}{c}{Synthetic Questions + Synthetic Answers}\\\hline
 None & 229297 & 79.1 & 87.9 & 78.2 & 86.8 \\
 Roundtrip (RT) & 93866 & 86.3 & 92.7 & 84.1 & 90.5  \\
 \textbf{Overgenerate \& RT} & \textbf{184992} & \textbf{87.1} & \textbf{93.2} & \textbf{85.2} & \textbf{91.5} \\
 \hline
Human Generated Data & 42472 & 86.3 & 93.2 & 82.4 & 89.7 \\
\hline
\end{tabular}}
\caption{\label{tab:qfilter_choices} Effect of filtration modeling choices on questions generated from ground truth and synthetic answers. 1.2 billion parameter models are used for every stage of the generation pipeline. Questions from no filtration are used in the other experiments with a second set of questions generated in overgeneration experiments.}
\vspace{-2mm}
\end{table}